\documentclass[times,twocolumn,final]{elsarticle}

\usepackage{arxiv}
\usepackage{tabularx}%
\usepackage{graphicx}%
\usepackage{multirow}%
\usepackage{amsmath,amssymb,amsfonts}%
\usepackage{amsthm}%
\usepackage{mathrsfs}%
\usepackage[title]{appendix}%
\usepackage[table]{xcolor}%
\usepackage{textcomp}%
\usepackage{manyfoot}%
\usepackage{booktabs}%
\usepackage{algorithm}%
\usepackage{algorithmicx}%
\usepackage{algpseudocode}%
\usepackage{listings}%
\usepackage{subcaption}%
\usepackage{tikz}%
\usetikzlibrary{calc,arrows.meta}%
\usepackage{pifont}%
\usepackage{multirow}%
\usepackage{tabularx}%
\usepackage{subcaption}%
\usepackage{pifont}%
\usepackage{adjustbox}
\usepackage{graphicx}
\usepackage{hyperref}
\usepackage{float}
\usepackage{placeins}
\usepackage{fancyhdr}
\usepackage{threeparttable}
\usepackage{epsfig, xspace, enumitem}
\usepackage{caption, overpic, textpos, pifont, adjustbox}
\newcounter{savedsubfigure}
\definecolor{rowgray}{gray}{0.98} 
\definecolor{extreme}{HTML}{ffa756}
\definecolor{majmin}{HTML}{d58a94}
\definecolor{iterpts}{HTML}{5170d7}
\definecolor{bbox}{HTML}{808080}
\definecolor{convexHull}{HTML}{50c864}
\colorlet{bboxlight}{bbox!40!white} 
\colorlet{iterptslight}{iterpts!65!white} 
\renewcommand{\headrulewidth}{0pt} 
\begin{document}
\fancypagestyle{firstpagestyle}{
    \fancyhf{} 
    \renewcommand{\headrulewidth}{0pt} 
    \fancyhead{} 
    \fancyfoot{} 
    \fancyhead[CO]{\em \fontsize{9pt}{8pt}\selectfont This article has been accepted to the International Conference on Information Processing in Computer-Assisted Interventions, 2026.}
}

\fancypagestyle{default}{
    \fancyhf{}
    \fancyhead[R]{\thepage} 
    \renewcommand{\headrulewidth}{0.4pt} 
    \fancyhead[LO]{\em \fontsize{9pt}{8pt}\selectfont This article has been accepted to the International Conference on Information Processing in Computer-Assisted Interventions, 2026.}
}


\title{S4M: 4-points to Segment Anything}

\author[1,2]{Adrien \snm{Meyer} \fnref{corresp}}
\fntext[corresp]{Corresponding author: \texttt{ameyer1@unistra.fr}}
\author[1,2]{Lorenzo \snm{Arboit}}
\author[1,2,3]{Giuseppe \snm{Massimiani}}
\author[2,4]{Shih-Min \snm{Yin}}
\author[1]{Didier \snm{Mutter}}
\author[1,2]{Nicolas \snm{Padoy}}

\address[1]{University of Strasbourg, CNRS, INSERM, ICube, UMR7357, Strasbourg, France}
\address[2]{IHU Strasbourg, France}
\address[3]{Fondazione Policlinico Universitario A. Gemelli IRCCS, Rome, Italy}
\address[4]{Department of General Surgery, Kaohsiung Chang Gung Memorial Hospital, Chang Gung University College of Medicine, Kaohsiung, Taiwan}

\received{XXX}
\finalform{XXX}
\accepted{XXX}
\availableonline{XXX}
\communicated{XXX}

\begin{abstract}
\noindent\textbf{Purpose:} 
The Segment Anything Model (SAM) promises to ease the annotation bottleneck in medical segmentation, but overlapping anatomy and blurred boundaries make its point prompts ambiguous, leading to cycles of manual refinement to achieve precise masks. Better prompting strategies are needed.

\noindent\textbf{Methods:}
To capture spatially meaningful cues at minimal annotation cost, we propose a structured prompting strategy using 4-points as a compact instance-level shape description. We study two 4-point variants: extreme points and the proposed major/minor axis endpoints, inspired by ultrasound measurement practice. Yet SAM cannot fully exploit such structured prompts, as it treats all points identically and lacks geometry-aware reasoning. To address this, we introduce S4M (4-points to Segment Anything), which augments SAM to interpret 4-points as relational cues rather than isolated clicks. S4M expands the prompt space with role-specific embeddings, disentangling each point’s semantic and spatial role. An auxiliary "Canvas" pretext task further strengthens prompt representation learning by sketching coarse masks directly from prompts, without visual input, fostering geometry-aware reasoning from minimal interaction.

\noindent\textbf{Results:} Across eight datasets in ultrasound and surgical endoscopy, S4M improves segmentation by +3.42 mIoU over a strong SAM baseline at equal prompt budget. An annotation study with three clinicians further demonstrate that major/minor prompts allow faster and practical annotation.

\noindent\textbf{Conclusion:} S4M increases performance, reduces annotation effort, and builds on conventions already embedded in clinical workflows. By aligning prompting with clinical practice, it removes a barrier to adoption and enables more scalable dataset development in medical imaging. We release our code and pretrained models at \href{https://github.com/CAMMA-public/S4M}{https://github.com/CAMMA-public/S4M}.
\\

\noindent\textbf{Keywords}: SAM, Extreme points, Major/Minor points, Ultrasound, Endoscopic
\end{abstract}

\maketitle
\thispagestyle{firstpagestyle}

\section{Introduction}\label{sec1}

Imagine an annotator clicking once on an intraoperative finding, only to see the response mask bleed into neighboring tissue. A second click repairs one edge but breaks another. This back-and-forth captures both the promise and frustration of interactive segmentation, where models generate masks from instance-specific prompts such as points or bounding boxes and refine them through user input. The Segment Anything Model~\cite{sam} (SAM) has popularized this paradigm, raising expectations that annotation could be scaled more efficiently and the dataset bottleneck eased, accelerating the development of medical computer vision applications such as landmark identification, skill assessment, and digital biopsy, which directly support clinical care.

\begin{figure*}[t]
    \centering
    \begin{subfigure}{\textwidth}
        \centering
            \begin{subfigure}{0.24\textwidth}
            \begin{tikzpicture}
              \node[inner sep=0,anchor=south west] (img) at (0,0)
                {\includegraphics[width=\linewidth]{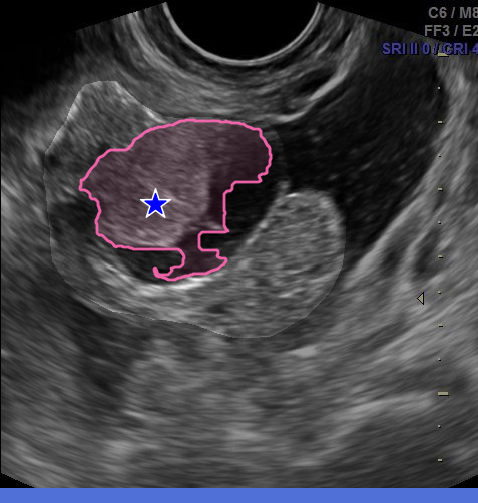}};
              \begin{scope}[x={(img.south east)},y={(img.north west)}]
                \draw[draw=white,line width=.9pt,double=white,double distance=1.5pt,line cap=round,
                      -{Triangle[length=2.5mm,width=3.2mm,fill=white]}]
                     (0.51,0.49) -- (0.38,0.57); 
              \end{scope}
            \end{tikzpicture}
            \end{subfigure}
            \begin{subfigure}{0.24\textwidth}
            \begin{tikzpicture}
              \node[inner sep=0,anchor=south west] (img) at (0,0)
                {\includegraphics[width=\linewidth]{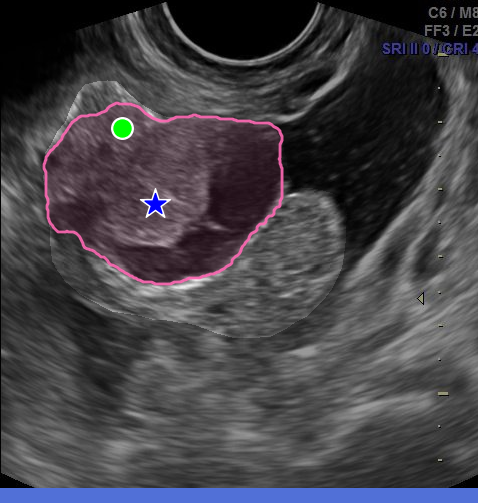}};
              \begin{scope}[x={(img.south east)},y={(img.north west)}]
                \draw[draw=white,line width=.9pt,double=white,double distance=1.5pt,line cap=round,
                      -{Triangle[length=2.5mm,width=3.2mm,fill=white]}]
                     (0.47,0.69) -- (0.31,0.72); 
              \end{scope}
            \end{tikzpicture}
            \end{subfigure}
            \begin{subfigure}{0.24\textwidth}
            \begin{tikzpicture}
              \node[inner sep=0,anchor=south west] (img) at (0,0)
                {\includegraphics[width=\linewidth]{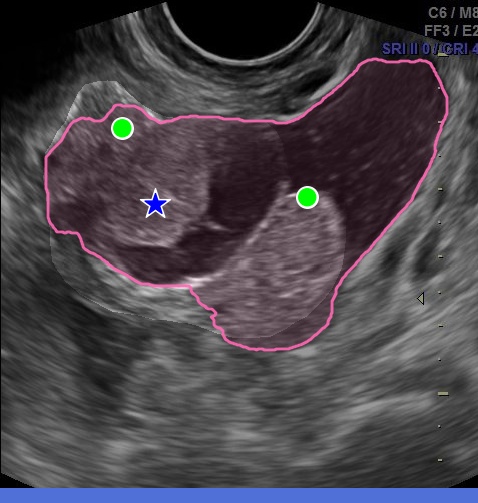}};
              \begin{scope}[x={(img.south east)},y={(img.north west)}]
                \draw[draw=white,line width=.9pt,double=white,double distance=1.5pt,line cap=round,
                      -{Triangle[length=2.5mm,width=3.2mm,fill=white]}]
                     (0.53,0.44) -- (0.62,0.57); 
              \end{scope}
            \end{tikzpicture}
            \end{subfigure}
            \begin{subfigure}{0.24\textwidth}
            \begin{tikzpicture}
              \node[inner sep=0,anchor=south west] (img) at (0,0)
                {\includegraphics[width=\linewidth]{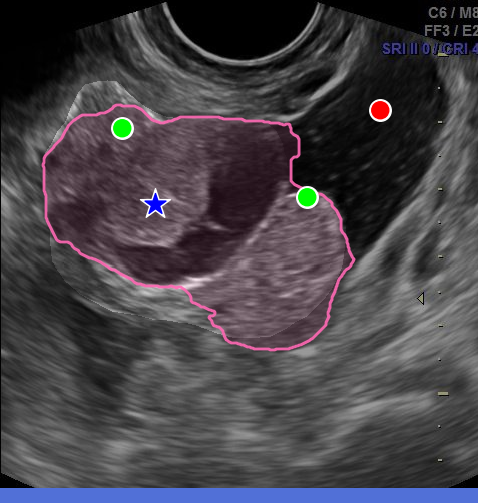}};
              \begin{scope}[x={(img.south east)},y={(img.north west)}]
                \draw[draw=white,line width=.9pt,double=white,double distance=1.5pt,line cap=round,
                      -{Triangle[length=2.5mm,width=3.2mm,fill=white]}]
                     (0.68,0.92) -- (0.78,0.81); 
              \end{scope}
            \end{tikzpicture}
            \end{subfigure}%
        \caption{Greedy region-based refinement loop with a specialist SAM (current workflow).}
    \label{subfig:greedy_region_based}
    \end{subfigure}
    \begin{subfigure}{\textwidth}
        \centering
        {%
            \setcounter{savedsubfigure}{\value{subfigure}}
            \setcounter{subfigure}{0}%
            \renewcommand{\thesubfigure}{\roman{subfigure}}
            \begin{subfigure}{0.24\textwidth}
                \includegraphics[width=\textwidth]{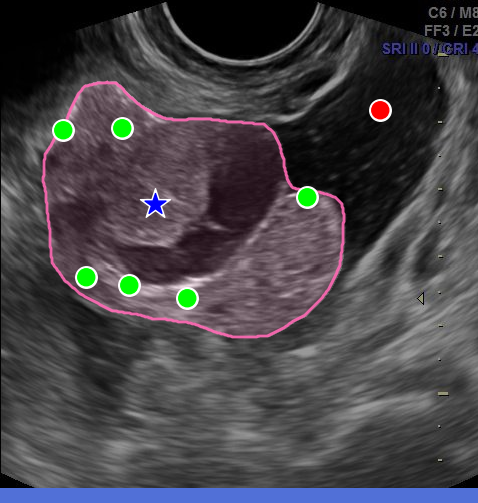}
                \caption{Region-based}
                \label{subfig:region_based}
            \end{subfigure}
            \begin{subfigure}{0.24\textwidth}
                \includegraphics[width=\textwidth]{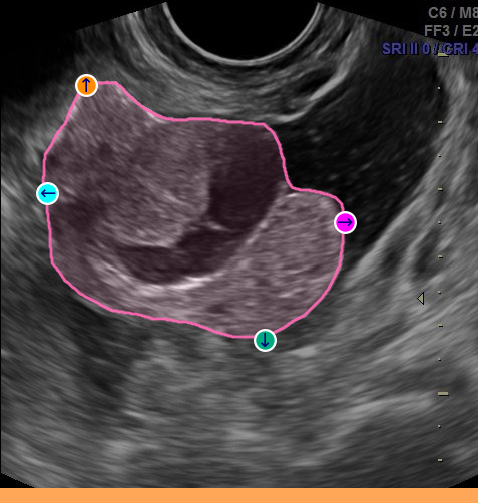}
                \caption{Extreme}
                \label{fig:extreme}
            \end{subfigure}
            \begin{subfigure}{0.24\textwidth}
            \begin{tikzpicture}
              \node[inner sep=0,anchor=south west] (img) at (0,0)
                {\includegraphics[width=\linewidth]{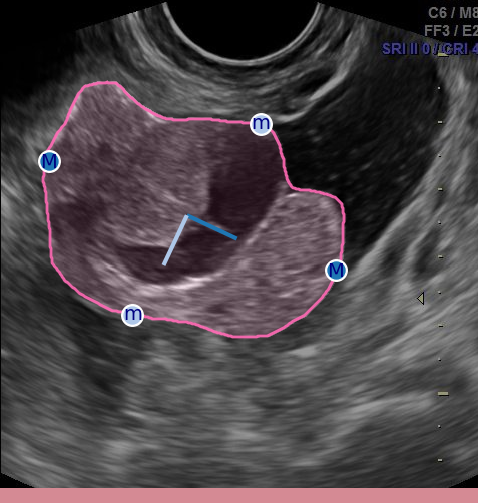}};
              \begin{scope}[x={(img.south east)},y={(img.north west)}]
                  \draw[draw={rgb,255:red,174;green,199;blue,232},line width=.9pt,double={rgb,255:red,174;green,199;blue,232},double distance=.15pt,line cap=round,
                  -{Triangle[length=1.5mm,width=1.5mm,fill={rgb,255:red,174;green,199;blue,232}]}]
                 (0.39,0.57) -- (0.34,0.46);
                \draw[draw={rgb,255:red,31;green,119;blue,180},line width=.9pt,double={rgb,255:red,31;green,119;blue,180},double distance=.15pt,line cap=round,
                      -{Triangle[length=1.5mm,width=1.5mm,fill={rgb,255:red,31;green,119;blue,180}]}]
                     (0.39,0.57) -- (0.505,0.52255);
              \end{scope}
            \end{tikzpicture}
            \caption{Major/Minor}
            \label{fig:majmin}
            \end{subfigure}
            \begin{subfigure}{0.24\textwidth}
                \includegraphics[width=\textwidth,trim=26 28 8 10,clip]{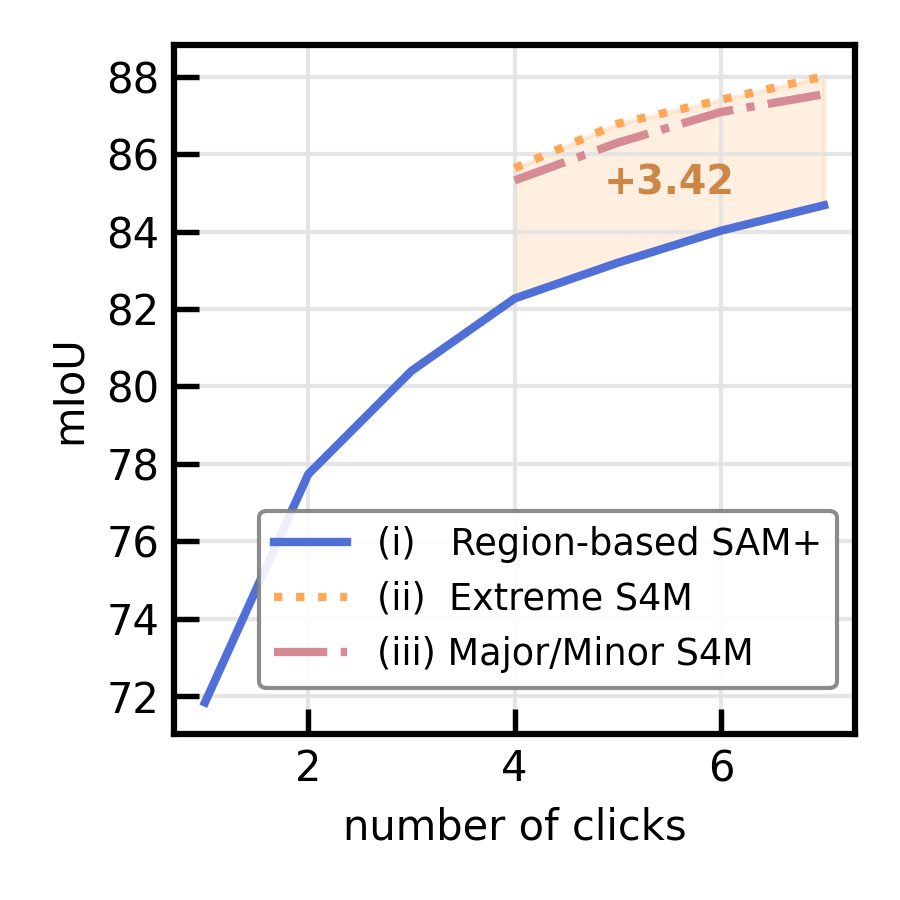}  
                \caption{mIoU vs \#points}
                \label{fig:mean_mIoU}
            \end{subfigure}
        }%
        \setcounter{subfigure}{\value{savedsubfigure}}
        \caption{Comparison of different prompt-placement strategies (ii and iii: proposed workflow).}
    \end{subfigure}
    \caption{\textbf{Prompt placement is key for efficient interactive segmentation.} In standard prompting (a), the initial click (blue star) is ambiguous, leading to iterative positive (green) and negative (red) refinement points. Our strategies instead \emph{directly} place prompts at Extreme-points (b.ii) or the proposed Major/Minor-points (b.iii, instance-oriented) to capture object shape, achieving higher mIoU with fewer interactions (b.iv), averaged over 8 datasets (6 supervised, 2 zero-shot).}
    \label{fig:abstract}
\end{figure*}

However, in the medical domain, overlapping structures, blurred boundaries, and heuristically defined contours undermine the specificity of point prompts, forcing annotators to rely on more descriptive prompts such as bounding boxes or numerous refinement points. In practice, point prompts trap users in a greedy correction loop: an initial click yields an imprecise mask, which is refined iteratively with new points, each patching local errors instead of working toward a coherent description of the instance as a whole (Fig.~\ref{subfig:greedy_region_based}). Worse, additional refinements can introduce new errors, as the instance identity is gradually established rather than defined upfront. Bounding boxes address identity more directly, but they only constrain the region of interest, leaving the model to infer boundaries with little guidance. They are also cognitively inefficient to draw, as annotators must rely on mental imagery to locate the object’s \emph{extremities} and position box corners that are not visually anchored, which slows annotation~\cite{papadopoulos2017extreme}.

A natural alternative is to mark these \emph{extremities} directly. The \emph{extreme points} method~\cite{papadopoulos2017extreme} asks annotators to click the top-, bottom-, left-, and right-most points of an instance (Fig.~\ref{fig:extreme}). This bypasses the need to manually place box corners, as they can be computed automatically, while also providing richer information through four boundary points anchored on the object itself. In their daily practice, clinicians rely on a related but distinct convention: to measure structures in ultrasound, they mark the endpoints of the major and minor axes, a simple gesture that encodes geometry, orientation, and extent (Fig.~\ref{fig:majmin}). We refer to these as the \emph{major/minor} points. Both \emph{4-points} strategies are fast, intuitive, and better capture instance identity, yet SAM-like models have not been adapted to exploit them. 

In this work, we revisit SAM through this clinical lens and introduce S4M (4-points to Segment Anything), designed to exploit both extreme and major/minor points as prompts. Each point type is assigned a dedicated learnable embedding, giving the model explicit awareness of their semantic roles and spatial relationships beyond generic positive points. To further strengthen this representation, we propose an auxiliary \emph{Canvas} training task that operates solely on prompts, without vision input. By predicting coarse masks from only 4-points, it encourages the model to learn the interdependence between them, how their spatial arrangement encodes shape, and to internalize this relationship for more robust segmentation.

Across eight datasets in ultrasound and surgical endoscopy, S4M achieves consistent gains of +3.42 mIoU over a strong specialist SAM baseline at equal prompt budget (Fig.~\ref{fig:mean_mIoU}). Importantly, it does so while reducing the cognitive load of annotation: by replacing iterative error-finding and correction with more informative 4-point prompts upfront, S4M makes the process faster and less demanding. As it modifies only the prompt encoding stage, the approach remains fully compatible with standard box, region-based, refinement prompts, and adaptable to other SAM-based designs. Finally, an annotation study with three clinicians demonstrates that major/minor point prompts can be collected efficiently, confirming the practical benefit of the proposed approach.

\begin{figure*}[t!]
    \centering

    \includegraphics[width=\textwidth]{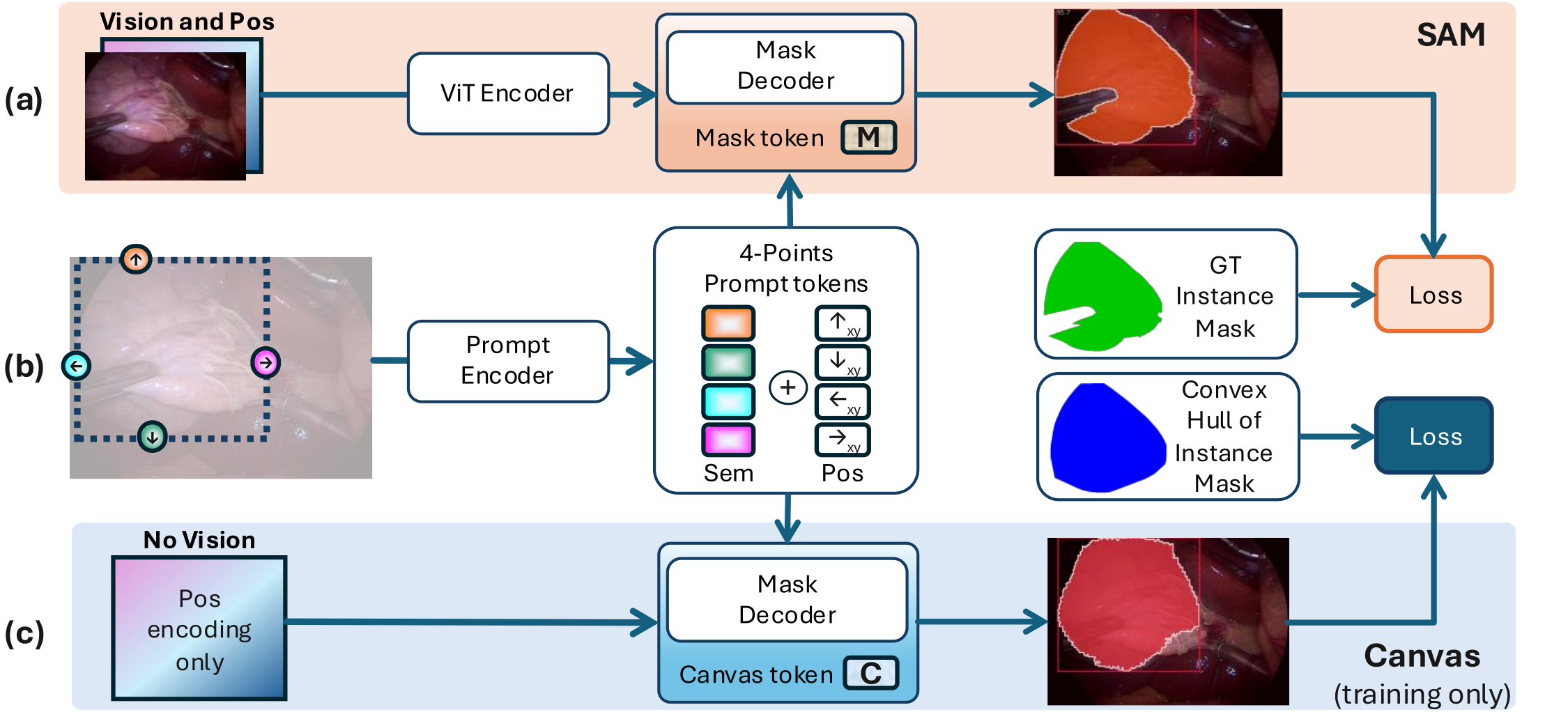}
\caption{\textbf{Overview of S4M.}
Built upon the standard SAM encoder/decoder (a), S4M introduces (b) role-aware 4-point prompts (extreme or major/minor) encoded with semantic and positional embeddings, and (c) an auxiliary Canvas task, a training-only decoder with its own Canvas token, supervised by the convex hull of the mask to strengthen 4-point representation learning. This example uses extreme points.}

    \label{fig:main_model}
\end{figure*}

\section{Background}\label{sec2}

\noindent \textbf{Segment Anything Model.}
SAM~\cite{sam} has redefined interactive segmentation, delivering impressive zero-shot performance through large-scale pretraining. It consists of three components: (1) an image encoder that computes a rich feature map of the input image, (2) a prompt encoder that embeds user inputs such as points or boxes using positional and learned embeddings, and (3) a lightweight mask decoder that combines these embeddings to generate segmentation masks. 
The decoder processes the encoded prompts together with learnable \emph{mask tokens} and a special \emph{IoU token}. Through alternating self-attention and cross-attention with the image embeddings, these tokens are iteratively updated. Each mask token is then mapped by an MLP into a dynamic linear classifier, applied to the upsampled image features via dot product to produce multiple candidate masks. The IoU token is used to predict a confidence score for each mask, allowing the model to estimate its own segmentation quality and select the best candidate automatically.
Thanks to this design, the decoder is efficient and can be rerun multiple times as users refine their prompts.

However, SAM is trained on natural images and transfers poorly to the medical domain. Recent adaptations such as MedSAM~\cite{medsam}, SAMed~\cite{sammed} and MedSA~\cite{medsa} reduce this gap by retraining on medical data, but the prompting design remains unchanged, and the inefficiency of point-based interaction persists.

\vspace{0.1\baselineskip}
\noindent \textbf{Extreme points.}
Extreme points~\cite{papadopoulos2017extreme} mark the top-, bottom-, left-, and right-most parts of an instance, placing clicks directly on its boundary from which the bounding-box can be derived easily. In natural images, extreme points have been shown to reduce annotation time compared to box annotation~\cite{papadopoulos2017extreme}. Deep Extreme Cut~\cite{maninis2018deep}, a CNN-based approach, further demonstrated that encoding extreme points as an additional input channel enhances interactive segmentation performance. In medical imaging, extreme points have been successfully leveraged for weakly supervised learning~\cite{roth2019weakly,roth2021going}  and confidence-guided segmentation~\cite{khan2019extreme}, demonstrating their efficiency in guiding model predictions with minimal annotation effort. Despite these advantages, extreme points have not been integrated into SAM-like interactive frameworks.

\section{Method}

S4M augments SAM~\cite{sam} to exploit 4-point prompts that encode object identity and shape more explicitly, bypassing early error-patching clicks and ensuring that every additional prompt refines rather than repairs. We consider two new prompting strategies for SAM-like approaches: \emph{extreme points}~\cite{papadopoulos2017extreme} and \emph{major/minor points}, the latter reflecting routine clinical measurement. We extend the prompt representation with role-aware embeddings to distinguish point types and introduce a \emph{Canvas} auxiliary task that teaches the model how their spatial arrangement encodes shape. Sharing the same architecture as SAM, S4M preserves its efficiency and compatibility with standard prompts. An overview of S4M is shown in Fig.~\ref{fig:main_model}.

\vspace{0.1\baselineskip}
\noindent \textbf{Prompt encoder.}
In SAM, each user prompt is represented by two embeddings: a positional encoding $\phi(x,y) \in \mathbb{R}^{256}$ that captures its spatial location, and a learned type embedding $e_t \in \mathbb{R}^{256}$ that specifies its semantic role (e.g., positive, negative, or bounding-box corner).
The set of learned type embeddings is:
\begin{equation}
E_{\text{SAM}} = \{e_{\text{pos}}, e_{\text{neg}}, e_{\text{boxA}}, e_{\text{boxB}}\},\quad  e_t \in E_{\text{SAM}}
\end{equation}
A prompt of type $t$ at coordinates $(x,y)$ is then encoded as
$p(x,y,t) = \phi(x,y) + e_t$.
This design is sufficient for standard region-based prompting, but becomes problematic when also supporting 4-points schemes. The same embedding $e_{\text{pos}}$ must denote a foreground region in one case, and a boundary anchor in the other. Optimizing for one role undermines the other, creating a representational mismatch
A simple solution might be to add one extra embedding for “extreme” prompts and one for “major/minor” prompts, allowing the model to at least distinguish them from generic positives. Yet this approach is still limited: it treats all 4-points within a scheme as interchangeable, discarding the geometric roles that make them informative.
S4M resolves this by introducing \emph{role-specific embeddings} that extend $E_{\text{SAM}}$:
\begin{equation}
E_{\text{S4M}} = E_{\text{SAM}} 
\cup \{{e_{\text{top}}, e_{\text{bottom}}, e_{\text{left}}, e_{\text{right}}}\} 
\cup \{{e_{\text{major}}, e_{\text{minor}}}\}
\end{equation}
By pairing each role with its position, the model learns not only \emph{where} a click lies and \emph{what} function it serves, but also how different points relate to one another. 
Taken together, these role-aware prompts implicitly convey an object’s extent and orientation, hinting at a coarse outline of its shape.

\begin{figure*}[t!]
    \centering
    \begin{subfigure}{0.24\textwidth}
        \begin{tikzpicture}
          \node[inner sep=0,anchor=south west] (img) at (0,0)
            {\includegraphics[width=\textwidth,trim=250 100 250 0,clip]{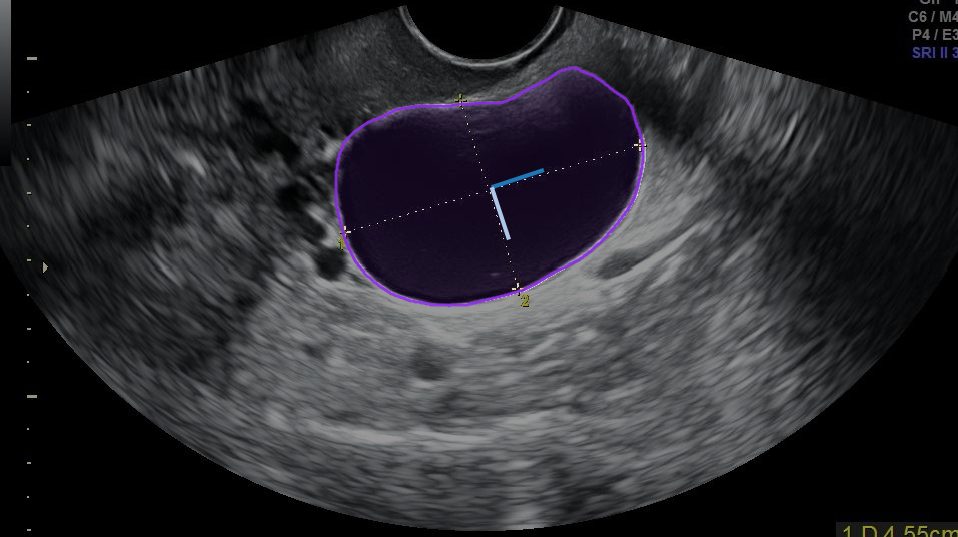}};
          \begin{scope}[x={(img.south east)},y={(img.north west)}]
              \draw[draw={rgb,255:red,174;green,199;blue,232},line width=.9pt,double={rgb,255:red,174;green,199;blue,232},double distance=.15pt,line cap=round,
              -{Triangle[length=1.5mm,width=1.5mm,fill={rgb,255:red,174;green,199;blue,232}]}]
             (0.53,0.565) -- (0.5675,0.445);
            \draw[draw={rgb,255:red,31;green,119;blue,180},line width=.9pt,double={rgb,255:red,31;green,119;blue,180},double distance=.15pt,line cap=round,
                  -{Triangle[length=1.5mm,width=1.5mm,fill={rgb,255:red,31;green,119;blue,180}]}]
                 (0.53,0.565) -- (0.66,0.62);
          \end{scope}
        \end{tikzpicture}
        \caption{PCA → axis}
        \label{subfig:prompt_pca_axis}
    \end{subfigure}
    \begin{subfigure}{0.24\textwidth}
        \begin{tikzpicture}
          \node[inner sep=0,anchor=south west] (img) at (0,0)
            {\includegraphics[width=\textwidth,trim=250 100 250 0,clip]{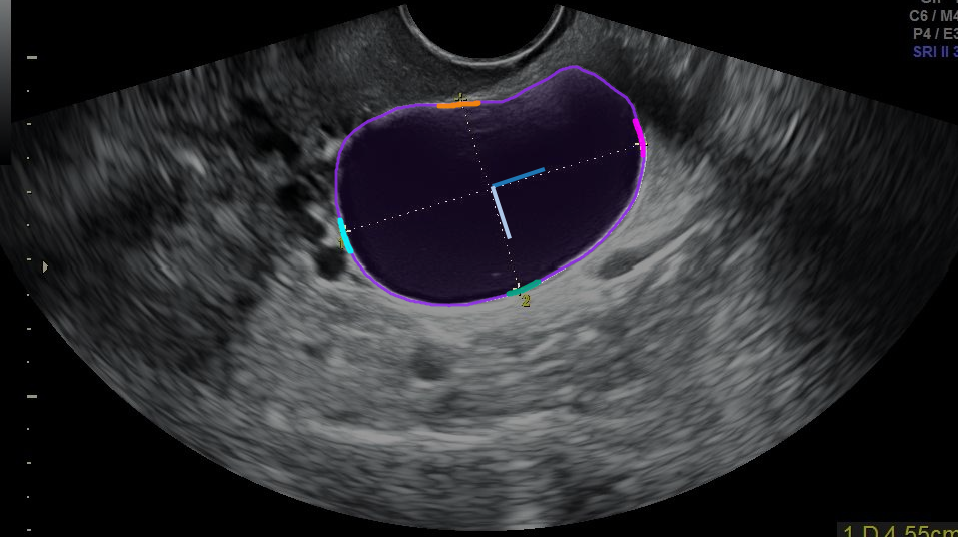}};
          \begin{scope}[x={(img.south east)},y={(img.north west)}]
              \draw[draw={rgb,255:red,174;green,199;blue,232},line width=.9pt,double={rgb,255:red,174;green,199;blue,232},double distance=.15pt,line cap=round,
              -{Triangle[length=1.5mm,width=1.5mm,fill={rgb,255:red,174;green,199;blue,232}]}]
             (0.53,0.565) -- (0.5675,0.445);
            \draw[draw={rgb,255:red,31;green,119;blue,180},line width=.9pt,double={rgb,255:red,31;green,119;blue,180},double distance=.15pt,line cap=round,
                  -{Triangle[length=1.5mm,width=1.5mm,fill={rgb,255:red,31;green,119;blue,180}]}]
                 (0.53,0.565) -- (0.66,0.62);
          \end{scope}
        \end{tikzpicture}
        \caption{Project pixels}
        \label{subfig:prompt_project}
    \end{subfigure}
    \begin{subfigure}{0.24\textwidth}
        \begin{tikzpicture}
          \node[inner sep=0,anchor=south west] (img) at (0,0)
            {\includegraphics[width=\textwidth,trim=250 100 250 0,clip]{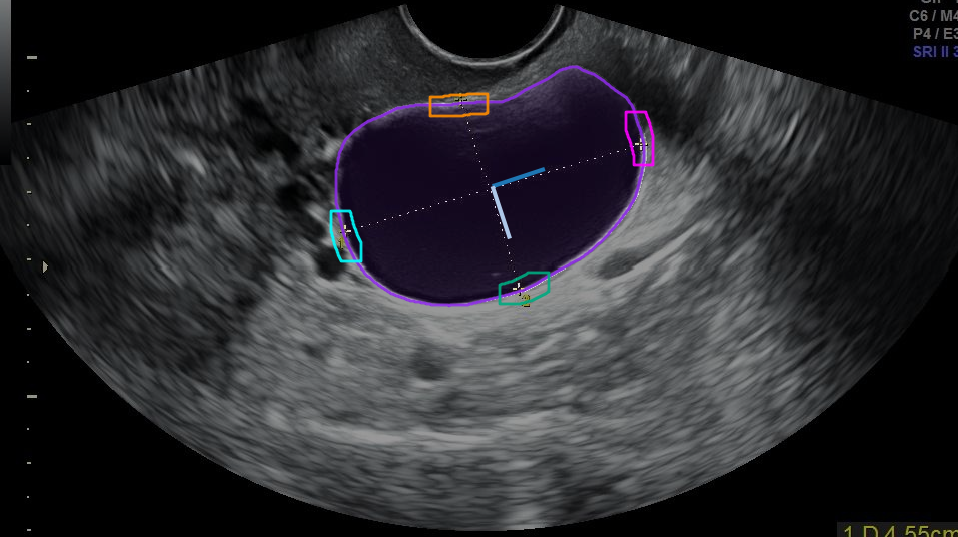}};
          \begin{scope}[x={(img.south east)},y={(img.north west)}]
              \draw[draw={rgb,255:red,174;green,199;blue,232},line width=.9pt,double={rgb,255:red,174;green,199;blue,232},double distance=.15pt,line cap=round,
              -{Triangle[length=1.5mm,width=1.5mm,fill={rgb,255:red,174;green,199;blue,232}]}]
             (0.53,0.565) -- (0.5675,0.445);
            \draw[draw={rgb,255:red,31;green,119;blue,180},line width=.9pt,double={rgb,255:red,31;green,119;blue,180},double distance=.15pt,line cap=round,
                  -{Triangle[length=1.5mm,width=1.5mm,fill={rgb,255:red,31;green,119;blue,180}]}]
                 (0.53,0.565) -- (0.66,0.62);
          \end{scope}
        \end{tikzpicture}
        \caption{Dilate → ROIs}
        \label{subfig:prompt_dilate}
    \end{subfigure}
    \begin{subfigure}{0.24\textwidth}
        \includegraphics[width=\textwidth,trim=250 100 250 0,clip]{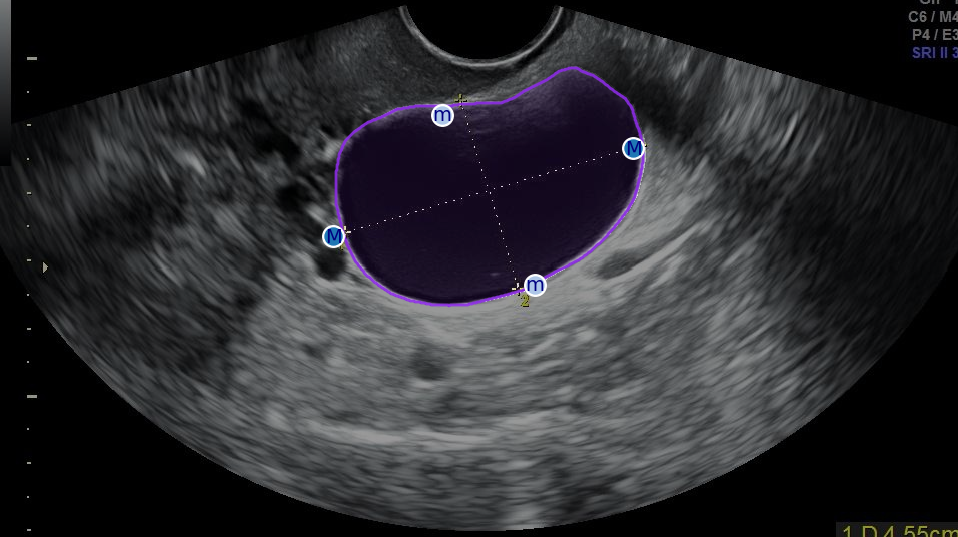}
        \caption{Sample}
        \label{subfig:prompt_sample}
    \end{subfigure}
    \caption{\textbf{4-points generation.} Our method emulates the measurements clinicians use in ultrasound, producing shape-aware 4-point prompts that correspond with the embedded measurement visible on the image (Zoom in for best view); for extreme points, the same procedure applies except that the image axis replaces the PCA axis.}
    
    \label{fig:prompt_gen}
\end{figure*}

\vspace{0.1\baselineskip}
\noindent \textbf{4-points generation.}
We simulate the initial 4-point prompts an annotator would provide by sampling them directly from the ground-truth masks (Fig.~\ref{fig:prompt_gen}). These prompts are generated either as major/minor points or as extreme points.

\vspace{0.1\baselineskip}
\noindent\emph{Major/minor points.}
We compute a two-component PCA on the border pixels to capture the dominant directions of shape variation (Fig.~\ref{subfig:prompt_pca_axis}).
The first and second principal axes are orthogonal by construction, and each is considered in both directions, giving four directions in total.
Border pixels are projected onto these directions (Fig.~\ref{subfig:prompt_project}), and candidate points $pts$ are ranked in each direction based on a score $s(pts)$:
\begin{equation}
s(pts) = w_{\text{main}}\,\pi_{\text{main}}(pts) - w_{\text{ortho}}\,\pi_{\text{ortho}}(pts)
\end{equation}
where $\pi_{\text{main}}(pts) $ and $\pi_{\text{ortho}}(pts) $ are normalized projections of $pts$ along the principal and orthogonal axes, weighted with $w_{\text{main}}$ and $w_{\text{ortho}}$.
We use $w_{\text{main}}=0.6$ and $w_{\text{ortho}}=0.4$ to favor endpoints that are extremal while remaining close to a perpendicular axis, consistent with the $\sim90^{\circ}$ axis placement clinicians use in practice.
The top-$k$ ranked pixels in each direction are then grouped into small \emph{top-ranked regions}, which are dilated into ROIs (Fig.~\ref{subfig:prompt_dilate}), from which final prompt positions are randomly sampled (Fig.~\ref{subfig:prompt_sample}).
This stochastic sampling keeps prompts \emph{near} the boundary, slightly inside or outside, capturing realistic annotator variability.

\vspace{0.1\baselineskip}
\noindent\emph{Extreme points.}
For extreme-point prompts, we reuse the same projection–ranking–sampling pipeline but replace PCA with the fixed horizontal and vertical image axes.
Here the orthogonality term is dropped, since the roles of top, bottom, left, and right are unambiguously defined by the image frame and do not rely on enforcing perpendicularity.
As with major/minor points, boundary-only pixels, ROI dilation, and random sampling ensure clicks remain realistic while introducing small variability around the object edges.

\vspace{0.1\baselineskip}
\noindent\emph{Refinement points.}
To mimic the iterative nature of interactive segmentation, we add two refinement rounds during training.
After the initial 4-points, each round samples one point uniformly from the current error region, which is then typed according to its location: \emph{positive} if inside a false negative area or \emph{negative} if inside a false positive area.
This process teaches the model to incorporate corrective clicks and progressively refine its predictions, as in the base SAM.

\vspace{0.1\baselineskip}
\noindent \textbf{Canvas.}
To strengthen the link between 4-points and shape, we introduce an auxiliary \emph{Canvas} training task that forces the model to reason from 4-points alone.
It mirrors SAM’s mask decoder but is trained with its own set of weights.
Instead of image embeddings, it receives a learned feature map of identical dimension summed with positional encodings.
With the same encoded prompts as the main decoder, the module must predict a coarse mask without any visual features.
Intuitively, this acts as a \emph{canvas} on which the model learns to sketch an object directly from 4-points, internalizing how their spatial arrangement encodes shape archetypes.

We use the convex hull of the ground-truth mask as supervision. 
This target strips away fine details and occlusions but preserves the global outline, aligning the prediction task with the limited information that 4-points can convey.
Because the prompt embeddings are shared across decoders, Canvas uses its own \emph{canvas token} in place of the regular mask token. This separates its predictions from the main task while still relying on the same 4-points representations.
The module is discarded at inference, serving purely as a training-time regularizer that consolidates the model’s understanding and representation learning of 4-point prompts.

\begin{table*}[t]
\centering
\caption{\textbf{Datasets} across two modalities: three for supervised train/validation/test, and one for zero-shot (ZS) testing. GIST: GastroIntestinal Stromal Tumors.}
\begin{tabular}{l l r r r l}
\toprule
\textbf{Modality} & \textbf{Dataset} & \textbf{Images} & \textbf{Instances} & \textbf{Classes} & \textbf{Procedure} \\
\midrule
\multirow{4}{*}{Ultrasound} 
& OTU        & 1,469  & 1,489   & 8  & Ovarian lesions \\
& \cellcolor{rowgray}BUSBRA       & \cellcolor{rowgray}1,875  & \cellcolor{rowgray}1,875   & \cellcolor{rowgray}2  & \cellcolor{rowgray}Breast lesions \\
& GIST-514        &   514  &   514   & 2  & GIST \\
& \cellcolor{rowgray}TN3K (ZS)    & \cellcolor{rowgray}3,493  & \cellcolor{rowgray}3,821   & \cellcolor{rowgray}1  & \cellcolor{rowgray}Thyroid nodules \\
\midrule
\multirow{4}{*}{Endoscopy} 
& Endoscapes-Seg50      &   493  & 2,248   & 6  & Cholecystectomy \\
& \cellcolor{rowgray}Hyper-Kvasir & \cellcolor{rowgray}1,000  & \cellcolor{rowgray}1,063   & \cellcolor{rowgray}1  & \cellcolor{rowgray}Colon polyps \\
& CaDIS           & 4,670  & 42,694  & 8  & Cataract surgery \\
& \cellcolor{rowgray}CholecSeg8k (ZS) & \cellcolor{rowgray}8,080  & \cellcolor{rowgray}48,146  & \cellcolor{rowgray}12 & \cellcolor{rowgray}Cholecystectomy \\
\bottomrule
\end{tabular}
\label{tab:datasets}
\end{table*}

\section{Experiments}

We evaluate S4M against a strong specialist SAM baseline (denoted SAM$^{+}$, see ablation for details) to test whether 4-points prompting translates into measurable gains in performance and efficiency. 
Experiments span two imaging modalities, ultrasound and surgical endoscopy, each with one model trained per modality for both S4M and SAM$^{+}$. 
Performance is compared across interaction budgets of 1–7 prompts, covering four regimes: standard region-based clicks with iterative refinements, bounding-box prompts, and our 4-points strategies with up to three refinements.

\vspace{0.1\baselineskip}
\noindent \textbf{Datasets.}
We benchmark on eight datasets across the two modalities (Table~\ref{tab:datasets}). 
For each modality, three datasets with official splits are used for supervised training, validation, and testing, while one dataset is held out entirely for zero-shot evaluation. 
We report mean Intersection-over-Union (mIoU), computed at the \emph{instance} level and averaged across classes.
To account for the stochasticity of prompt sampling, each experiment is repeated five times with independently generated prompts, and we report mean and standard deviation across runs.

\begin{figure*}[t]\centering
\includegraphics[width=\textwidth,trim=10 33 10 10,clip]{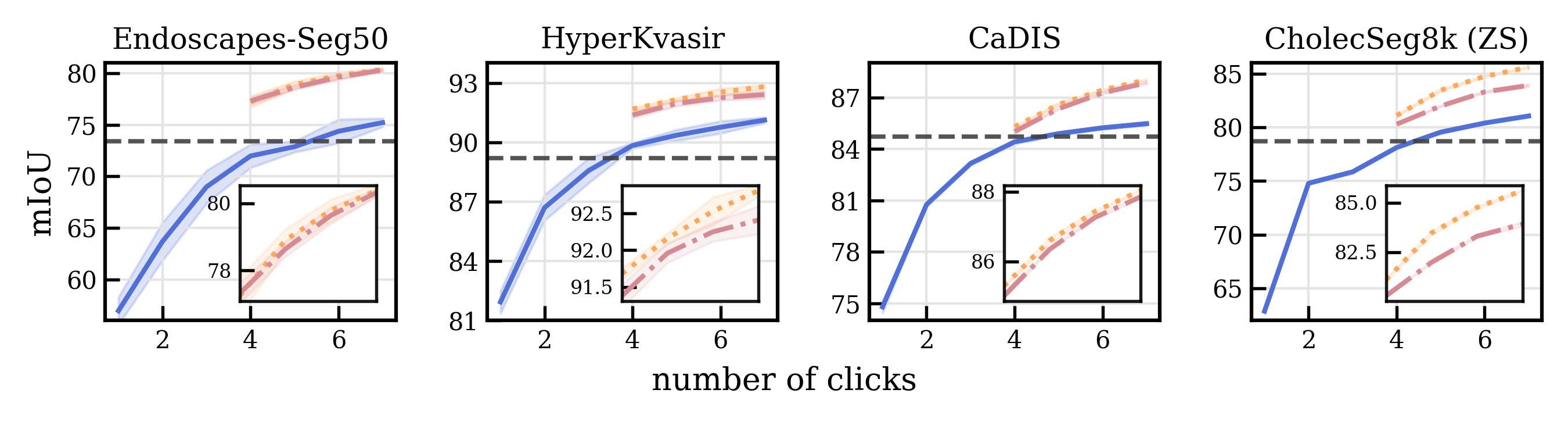}
\includegraphics[width=\textwidth,trim=10 15 10 2,clip]{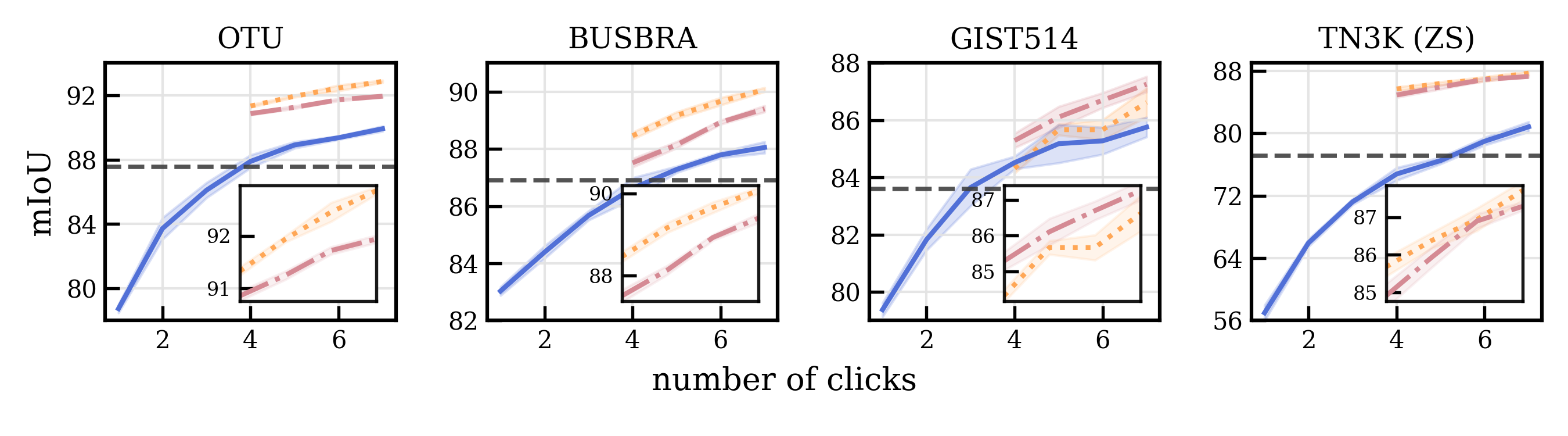}
\caption{\textbf{Evolution of performance with different prompt-budget.} Per-dataset mean IoU comparison for surgical endoscopy (top row) and ultrasound (bottom row) under supervised (first three columns) and zero-shot (ZS, last column) settings. Across all datasets, both \colorbox{extreme}{\strut \textbf{extreme}} (orange dotted) and \colorbox{majmin}{\strut \textbf{major/minor}} points (pink dash-dotted) using S4M outperform both \colorbox{iterptslight}{\strut \textbf{region-based}} points (blue solid) and \colorbox{bboxlight}{\strut \textbf{bounding box}} (dashed gray line) SAM$^{+}$ baselines.}
\label{fig:main_results}
\end{figure*}

\vspace{0.1\baselineskip}
\noindent \textbf{Implementation details}
We initialize SAM and the Canvas module using the pretrained SAM ViT-b model and then finetune on one H100 GPU for 1600 iterations with a batch size of 6. Images are resized and then padded to 1024x1024, maintaining aspect ratio. Our code is based on MMDetection v3.3~\cite{mmdet}. We use the AdamW optimizer, with an initial learning rate of $1 \times 10^{-4}$ and a warm-up period of 500 iterations. We reduce the learning rate by a factor of 10 at 1000 and 1400 iterations. We train SAM$^{+}$/S4M by sampling the prompt type (points, box, extreme or major/minor points) uniformly. Following SAM~\cite{sam}, we use a linear combination of focal and dice loss (20:1) for both the mask decoder and the canvas task.

\section{Results}
We present quantitative results in Fig.~\ref{fig:main_results}. S4M consistently outperformed the specialist SAM$^{+}$ across all datasets and interaction budgets, with an average gain of +3.42 mIoU that held in both supervised and zero-shot settings. Importantly, results are also more stable, as evidenced by lower standard deviation across runs with different prompts. Both extreme and major/minor prompts achieved higher mIoU with fewer clicks, while region-based SAM$^{+}$ showed steep gains in the first refinements but plateaued quickly. In surgical endoscopy, the first three refinements improved performance by $+10.85$ mIoU over the initial click ($+5.96$ in ultrasound), whereas the next three refinements added only marginal gains ($+1.86$ and $+1.57$, respectively), underscoring diminishing return. Since multiple clicks are needed in practice to reach acceptable performance, S4M leverages the same budget upfront in a structured 4-points scheme, yielding superior performance without iterative re-runs and sparing annotators the cognitive burden of analyzing intermediate outputs. Both 4-point methods performed comparably, while major/minor points were quicker to annotate (See section~\ref{section:ann_study}).

We present qualitative results in Fig.~\ref{fig:quali} and in the supplementary material. The Canvas proto-masks, generated from prompts alone without visual input, demonstrate that S4M has internalized how 4-points encode global shape, thereby improving the final masks. The resulting segmentations capture complex, highly concave structures more faithfully, with the 4-points anchoring multiple parts of the object boundary.

\begin{figure*}[t]
\centering

\begin{tikzpicture}
\node[anchor=south west,inner sep=0] (img1) {\includegraphics[width=0.49\linewidth]{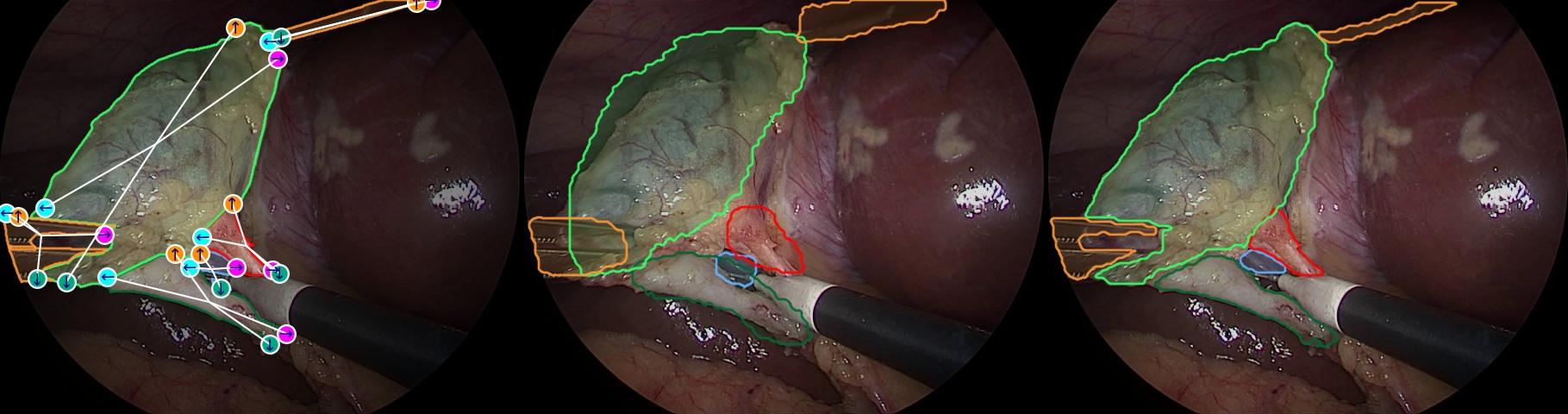}};
\begin{scope}[x={(img1.south east)},y={(img1.north west)}]
  \node[text=white, font=\bfseries, align=center] at (0.17,0.10) {Prompts};
  \node[text=white, font=\bfseries, align=center] at (0.50,0.11) {Canvas};
  \node[text=white, font=\bfseries, align=center] at (0.83,0.11) {Prediction};
\end{scope}
\end{tikzpicture}
\hfill
\begin{tikzpicture}
\node[anchor=south west,inner sep=0] (img2) {\includegraphics[width=0.49\linewidth, trim=0 142 0 0, clip]{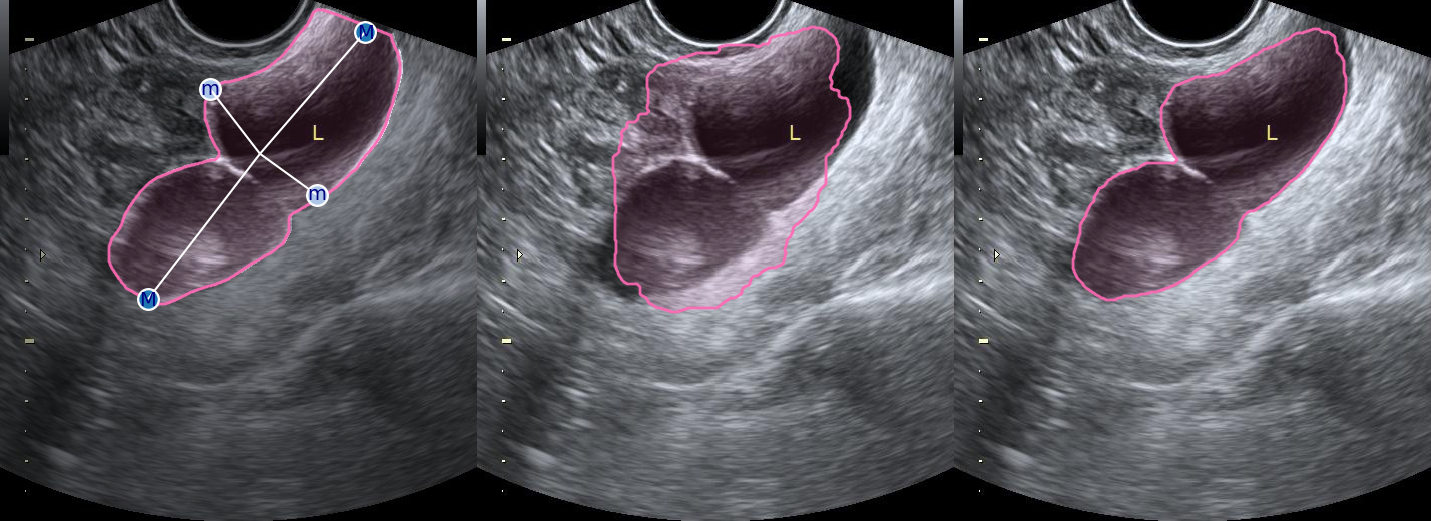}};
\begin{scope}[x={(img2.south east)},y={(img2.north west)}]
  \node[text=white, font=\bfseries, align=center] at (0.17,0.11) {Prompts};
  \node[text=white, font=\bfseries, align=center] at (0.50,0.11) {Canvas};
  \node[text=white, font=\bfseries, align=center] at (0.83,0.11) {Prediction};
\end{scope}
\end{tikzpicture}

\caption{S4M results on Endoscapes (left, extreme) and OTU (right, major/minor).}
\label{fig:quali}
\end{figure*}

\begin{figure*}[t]
    \centering
    \begin{subfigure}{0.32\textwidth}
      \raisebox{0.9mm}[0pt][0.9mm]{\includegraphics[width=\textwidth]{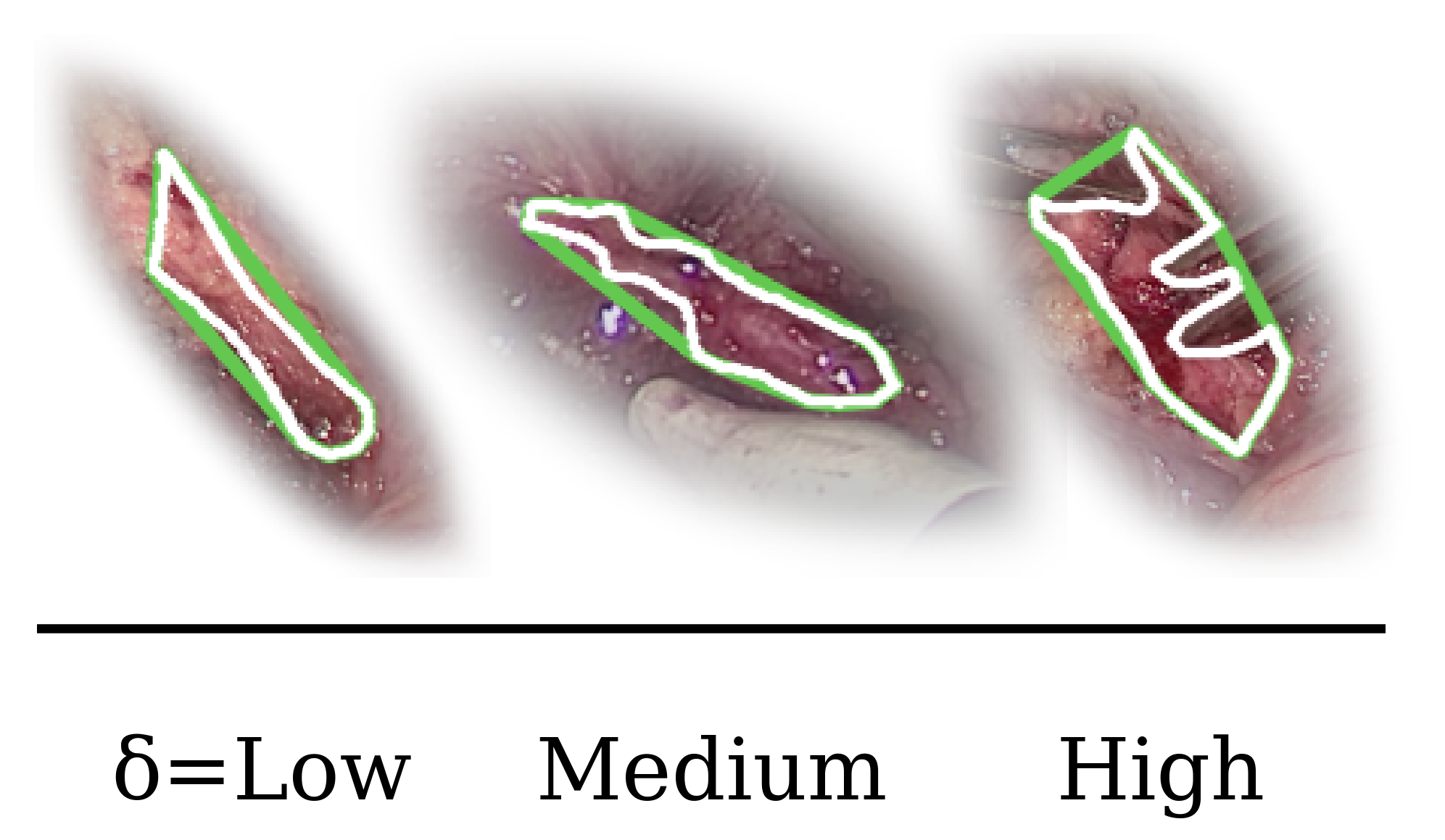}}
      \caption{Examples of Concavity}
      \label{subfig:example_concavity}
    \end{subfigure}
    \begin{subfigure}{0.33\textwidth}
        \includegraphics[width=\textwidth]{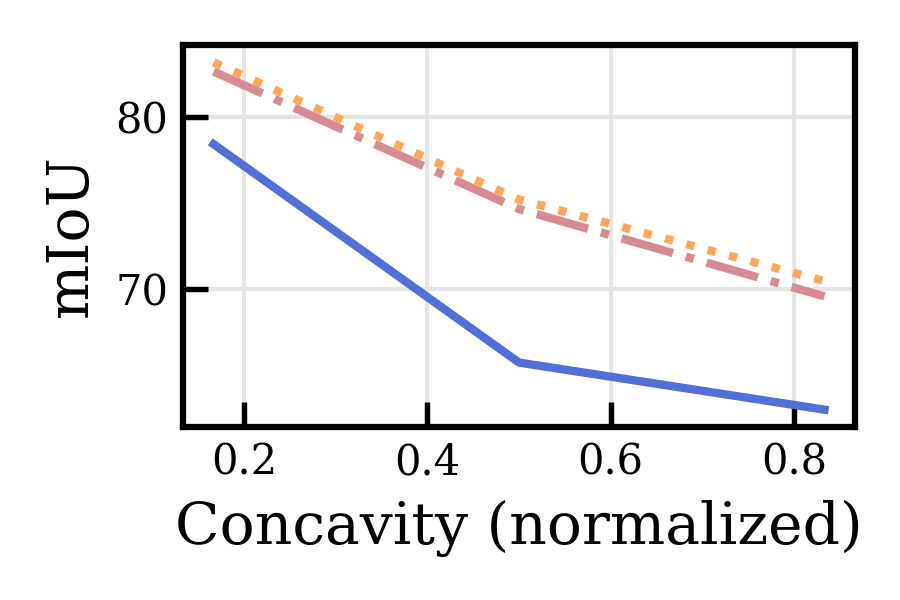}
        \caption{Endoscapes}
        \label{subfig:concavity_endoscape}
    \end{subfigure}
    \begin{subfigure}{0.33\textwidth}
        \includegraphics[width=\textwidth]{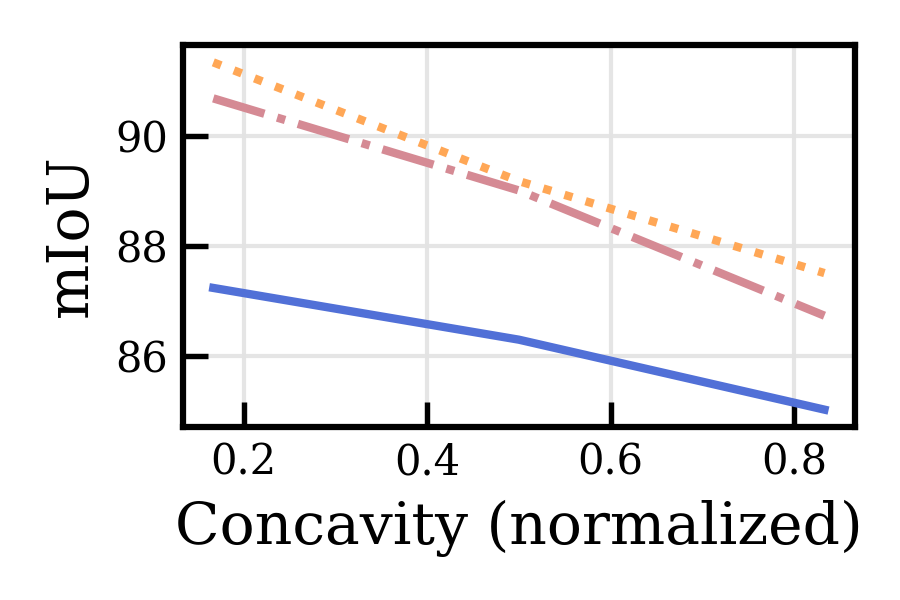}
        \caption{OTU}
        \label{subfig:concavity_MMOTU}
    \end{subfigure}
    \caption{\textbf{Effect of shape concavity on segmentation.} 
    (a) Examples of structures with increasing concavity $\delta$ (\colorbox{convexHull}{\strut \textbf{convex hull}} in green).  
    (b–c) mIoU stratified by concavity in Endoscapes and OTU.  
    While performance drops as shapes become more irregular, S4M with \colorbox{extreme}{\strut \textbf{extreme}} or \colorbox{majmin}{\strut \textbf{major/minor}} prompts consistently preserves higher mIoU, showing greater robustness than \colorbox{iterptslight}{\strut \textbf{region-based}} (4-points) SAM$^{+}$.}    
    \label{fig:concavity}
\end{figure*}

\vspace{0.1\baselineskip}
\noindent \textbf{Shape Complexity.}
Not all structures are equally easy to segment. Simple, convex shapes can often be captured with minimal prompting, but highly concave or irregular ones remain difficult, as region- and box-based interaction provides little geometric constraint. These challenging shapes are common in medical imaging and drive up annotation cost, making robustness to geometric complexity a critical test for any interactive framework.
We quantify complexity with a concavity index $\delta = 1 - \frac{\text{Area(mask)}}{\text{Area(convex hull)}}$, normalized per dataset. Values near $0$ correspond to convex, regular shapes, while higher values indicate irregular and deeply indented outlines (examples in Fig.~\ref{subfig:example_concavity}). Fig.~\ref{fig:concavity} shows performance stratified by concavity. Across both ultrasound and endoscopy, performance decreases as concavity grows, confirming the added difficulty of complex shapes. Yet S4M consistently narrows this gap: 4-point prompts maintain higher mIoU at all levels of concavity, with less performance drop in the higher concavity regimes. This indicates that explicitly encoding object geometry through four structured points makes the model more resilient to irregular boundaries than region-based prompting alone.

\begin{figure*}[t]\centering
\begin{subfigure}{0.43\textwidth}
    \includegraphics[width=\textwidth,trim=0 12 260 10,clip]{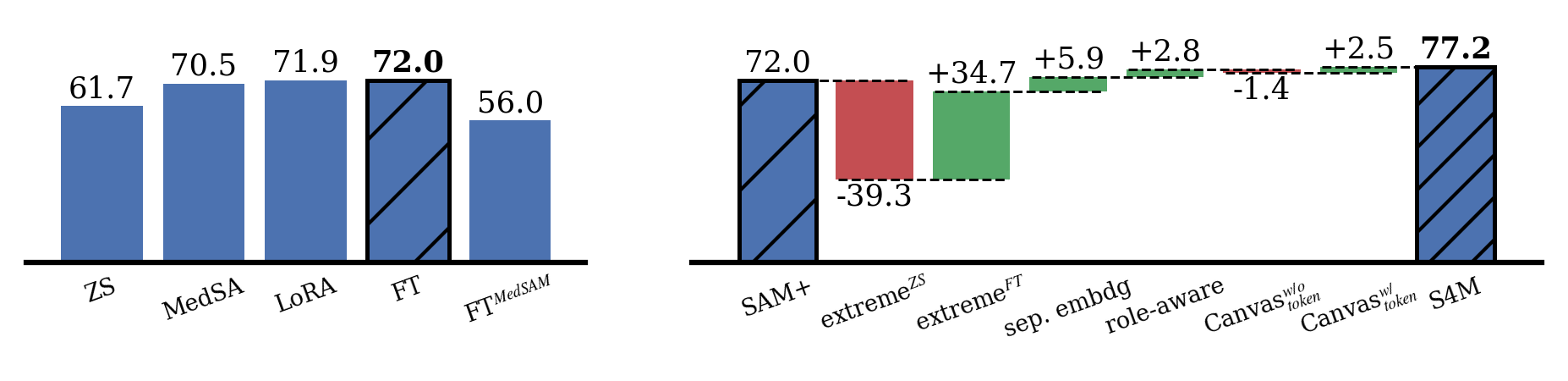}
    \caption{Finetuning choice for SAM$^{+}$} 
    \label{ablation:SAM$^{+}$_choice}
\end{subfigure}
\begin{subfigure}{0.54\textwidth}
    \includegraphics[width=\textwidth,trim=200 12 10 10,clip]{ablation_barplots.png}
    \caption{Impact of Components} 
    \label{ablation:SAM$^{+}$_to_S4M}
\end{subfigure}
\caption{\textbf{Ablation results.} Model selection and component analysis for SAM$^{+}$ and S4M (Endoscapes). \textit{Sep. embdg}: separates embeddings for positive and extreme points.}
\label{fig:ablation_barplot}
\end{figure*}

\begin{figure*}[t]\centering
\includegraphics[width=\textwidth]{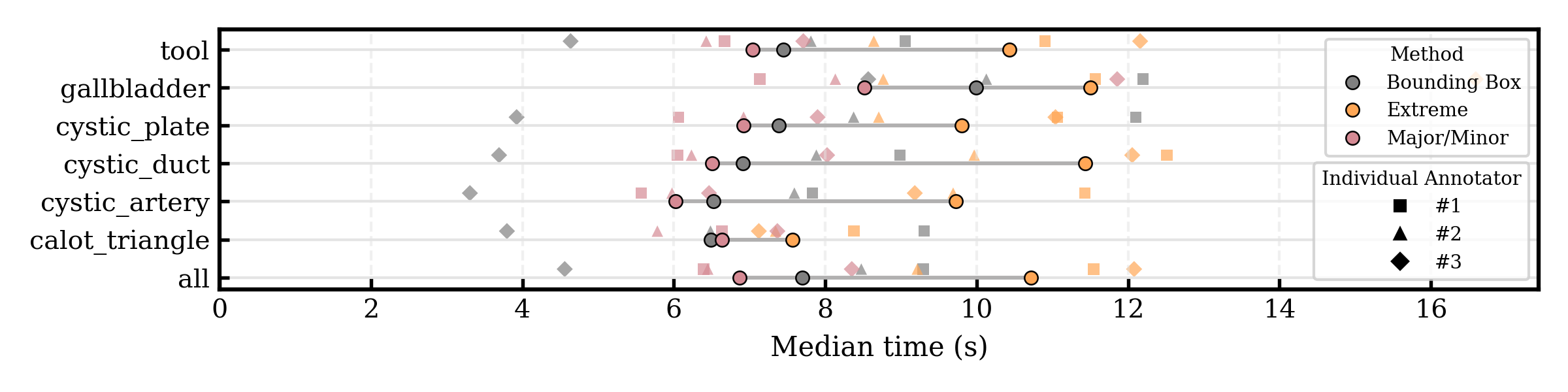}
\caption{Annotation time comparison on the Endoscapes-Seg50 test set.}
\label{fig:annotation_study}
\end{figure*}

\vspace{0.1\baselineskip}
\noindent \textbf{SAM$^{+}$ baseline selection.}
We conduct an ablation to identify the optimal strategy for training our SAM$^{+}$ baseline, which our method builds upon with the proposed modules. As shown in Fig.~\ref{ablation:SAM$^{+}$_choice}, we evaluate three adaptation methods for SAM~\cite{sam} and MedSAM~\cite{medsam} on Endoscapes: Medical SAM-adapter~\cite{medsa} (MedSA), Low-Rank Adaptation~\cite{lora} (LoRA), and end-to-end finetuning (FT). All settings use the same data, augmentations, and training schedule. LoRA and FT yield comparable results and both outperform MedSA, while starting from the original SAM weights proves more effective than from MedSAM. We therefore adopt FT as our training strategy, defining the resulting fully fine-tuned model as our SAM$^{+}$ baseline, on which S4M is built.

\begin{table}[t]
\caption{\textbf{Prompt evaluation on Endoscapes (mIoU).} (left) Oracle vs.\ IoU-score mask refinement. (right) Across prompt types; SAM$^{+}_{FT}$ uses 4-points training.}
\begin{tabular}{@{}cc@{}}
\begin{subtable}[t]{0.42\linewidth}
\centering
\resizebox{\linewidth}{!}{
\begin{tabular}{lcccc}
\toprule
\multirow{2}{*}{\textbf{Method}} & \multicolumn{4}{c}{\textbf{Number of Points}} \\
\cmidrule(lr){2-5}
 & \textbf{1} & \textbf{2} & \textbf{3} & \textbf{4} \\
\midrule
IoU token & 56.9 & 63.7 & 68.9 & 72.0 \\
\cellcolor{rowgray}oracle & \cellcolor{rowgray}58.0 & \cellcolor{rowgray}64.3 & \cellcolor{rowgray}69.2 & \cellcolor{rowgray}72.1 \\
extreme & - & - & - & 77.2 \\
\bottomrule
\end{tabular}}
\end{subtable}
&
\begin{subtable}[t]{0.55\linewidth}
\centering
\resizebox{\linewidth}{!}{
\begin{tabular}{lcccc}
\toprule
\multirow{2}{*}{\textbf{Method}} & \multicolumn{4}{c}{\textbf{Prompt types}} \\
\cmidrule(lr){2-5}
& \textbf{4-Region} & \textbf{Box} & \textbf{extr.} & \textbf{maj/min} \\
\midrule
SAM$^{+}$ & 72.0 & 73.4 & 32.7 & 32.5\\
\cellcolor{rowgray}SAM$^{+}_{FT}$ & \cellcolor{rowgray}67.0 & \cellcolor{rowgray}73.3 & \cellcolor{rowgray}67.4 &  \cellcolor{rowgray}67.6\\
S4M & 71.8 & 73.6 & 77.2 & 77.3 \\
\bottomrule
\end{tabular}}
\end{subtable}
\\ 
\end{tabular}
\label{tab:results_combined}
\end{table}

\vspace{0.1\baselineskip}
\noindent \textbf{S4M Ablation Study.}
We ablate the components of S4M to measure their effect (Fig.~\ref{ablation:SAM$^{+}$_to_S4M}).
When tested with unseen extreme points, the SAM$^{+}$ baseline drops sharply, revealing weak generalization to new prompt types.
Adding extreme prompts to training reduces this gap but remains suboptimal: using a single embedding for both positive and extreme points creates representational conflict.
Separating these embeddings eliminates the interference and restores performance, while role-aware encoding further boosts performance by teaching the model to interpret 4-points as a coherent geometric structure.
Finally, the Canvas task harms results when added without its own token but, once decoupled with a dedicated Canvas token, increases performance. Overall, S4M achieves a \textbf{+5.2 mIoU} improvement on Endoscapes compare to SAM$^{+}$.

\vspace{0.1\baselineskip}
\noindent \textbf{Oracle SAM$^{+}$.}
SAM outputs multiple masks and selects one via its IoU token, which is not always optimal. To estimate an upper bound, we use an \emph{oracle} variant that selects, at each step, the mask with highest IoU to the ground truth, simulating a human choosing the best candidate and sampling the next point from the remaining error.
On Endoscapes (Table~\ref{tab:results_combined}, left), the oracle yields only marginal gains over the IoU token selection, with diminishing gains as more refinements are added as ambiguity decreases. Even so, it remains 5.1 mIoU below S4M at 4-points, despite the extra annotation cost of manual mask selection.

\vspace{0.1\baselineskip}
\noindent \textbf{Prompt Generalization.}
Table~\ref{tab:results_combined} (right) shows that S4M matches SAM$^{+}$ on region and box prompts, while achieving the best results on both 4-points variants (77.2 and 77.3 mIoU for extreme and major/minor). By contrast, SAM$^{+}_{FT}$, fine-tuned with 4-point prompts, narrows the gap on these settings (extreme: 67.4, major/minor: 67.6) but at the cost of lower performance on region-based prompts (67.0 vs.\ 72.0). This suggests S4M’s role-aware design avoids prompt-type interference, delivering high performance without compromising other interaction modes.

\vspace{0.1\baselineskip}
\noindent \textbf{Annotation study.}\label{section:ann_study}
To assess annotation efficiency, we conducted a study with three clinicians on the Endoscapes test set. Each participant annotated the same structures using boxes, extreme, and major/minor points in separate sessions. As shown in Fig.~\ref{fig:annotation_study} and contrary to findings in natural images~\cite{papadopoulos2017extreme}, boxes were faster than extreme points. Major/minor points, however, were the quickest to collect, though inter-annotator variability remained. Annotator \#3, for instance, was faster with boxes. This suggests that efficiency depends on user habits, highlighting the need for models that support multiple prompt types. Despite the sample size, these results indicate that major/minor points offer a promising trade-off between speed and segmentation quality.


\section{Conclusion}
S4M addresses a limitation overlooked by previous SAM variants: the prompt design itself. By shifting from reactive correction to proactive geometric description, it treats four structured points as a compact yet powerful cue to object shape. Role-aware embeddings and the Canvas task teach the model to reason over spatial relations rather than correct local errors, delivering consistent performance gains and faster annotation. Beyond performance, this reframing aligns interaction with clinical measurement habits and moves interactive segmentation toward more scalable and human-aligned segmentation models in medical imaging.

\vspace{0.1\baselineskip}
\noindent \textbf{Acknowledgements.}
This work was supported by French state funds managed within the ’Plan Investissements d’Avenir’ funded by the ANR under references ANR-21-RHUS-0001 (DELIVER) and ANR-10-IAHU-02 (IHU Strasbourg). This work was performed using HPC resources managed by CAMMA, IHU Strasbourg, Unistra Mesocentre, and GENCI-IDRIS (Grant AD011013710R3).

\noindent \textbf{Disclosure of potential conflicts of interest.}
The authors declare no conflict.

\bibliographystyle{sn-basic}
\bibliography{sn-bibliography}

\end{document}